\def\year{2021}\relax
\documentclass[letterpaper]{article} 
\usepackage{aaai21}  
\usepackage{times}  
\usepackage{helvet} 
\usepackage{courier}  
\usepackage[hyphens]{url}  
\usepackage{graphicx} 
\usepackage{natbib}  
\usepackage{caption} 
\usepackage{hyperref}
\usepackage{subcaption}
\usepackage{xcolor}
\usepackage{amsmath}
\usepackage{cleveref}
\usepackage{algorithmic}
\usepackage{algorithm}
\usepackage{multirow}
\usepackage{xcolor}
\usepackage[switch]{lineno}

\newcommand{\mat}[1]{\mathbf{#1}}
\renewcommand{\vec}[1]{\mathbf{#1}}
\let\oldhat\hat
\renewcommand{\hat}[1]{\oldhat{\mathbf{#1}}}

\title{LIREx: Augmenting Language Inference with Relevant Explanation}
\author{
    Xinyan Zhao,\textsuperscript{\rm 1}
    V.G.Vinod Vydiswaran\textsuperscript{\rm 2,1}\\
}
\affiliations{
    \textsuperscript{\rm 1}School of Information;
    \textsuperscript{\rm 2}Department of Learning Health Sciences \\
    University of Michigan, Ann Arbor, Michigan 48109 USA\\
    \{zhaoxy, vgvinodv\}@umich.edu

}

\begin{document}
\maketitle

\begin{abstract}
Natural language explanations (NLEs) are a special form of data annotation in which annotators identify rationales (most significant text tokens) when assigning labels to data instances, and write out explanations for the labels in natural language based on the rationales. 
NLEs have been shown to capture human reasoning better, but not as beneficial for natural language inference (NLI). In this paper, we analyze two primary flaws in the way NLEs are currently used to train explanation generators for language inference tasks. We find that the explanation generators do not take into account the variability inherent in human explanation of labels, and that the current explanation generation models generate spurious explanations. To overcome these limitations, we propose a novel framework, LIREx, that incorporates both a rationale-enabled explanation generator and an instance selector to select only relevant, plausible NLEs to augment NLI models. When evaluated on the standardized SNLI data set, LIREx achieved an accuracy of 91.87\%, an improvement of 0.32 over the baseline and matching the best-reported performance on the data set. It also achieves significantly better performance than previous studies when transferred to the out-of-domain MultiNLI data set. Qualitative analysis shows that LIREx generates flexible, faithful, and relevant NLEs that allow the model to be more robust to spurious explanations. The code is available at~\url{https://github.com/zhaoxy92/LIREx}.

\end{abstract}

\section{Introduction}
\label{sec:intro}
Natural language explanations (NLEs) provided at the time of assigning labels to data instances are likely to better capture human reasoning than the labels alone. NLEs are special forms of data annotation in which annotators identify both the class label and the rationales (most significant text tokens), and write out explanations in natural language. They have been suggested to potentially improve the performance and interpretability of deep learning-based models -- i.e.~either augmenting model performance by incorporating NLEs as additional contextual features, or explaining model decisions by training an explanation generator.
Researchers have parsed NLEs into structured logical forms~\cite{srivastava2017joint,hancock2018training,lee2020lean,qin2020learning} or directly encoded them into a vector-based semantic representation~\cite{fidler2017teaching}. 
Recent success in language modeling and generation have enabled trained models to explicitly provide human-readable explanations for classification tasks~\cite{kim2018textual,huk2018multimodal,camburu2018snli,kumar2020nile,rajani2019explain}. Similarly, studies such as \cite{rajani2019explain} have reported significant performance improvements on commonsense reasoning tasks by including NLEs in training a language generation model. However, these trends do not carry over to natural language inference (NLI) task in which a premise-hypothesis pair is expected to be classified into \textit{entailment}, \textit{neutral}, or \textit{contradiction}. Previous studies on utilizing NLEs for NLI tasks have reported a drop in overall performance, even with powerful deep learning-based models such as LSTM~\cite{camburu2018snli}, RoBERTa, and GPT2~\cite{kumar2020nile}. We study this discrepancy in more detail and identify two primary issues, described below, with how NLEs have been incorporated for the NLI task so far. 
\begin{figure}
     \centering
     \begin{subfigure}[b]{0.5\textwidth}
         \centering
         \includegraphics[width=\textwidth]{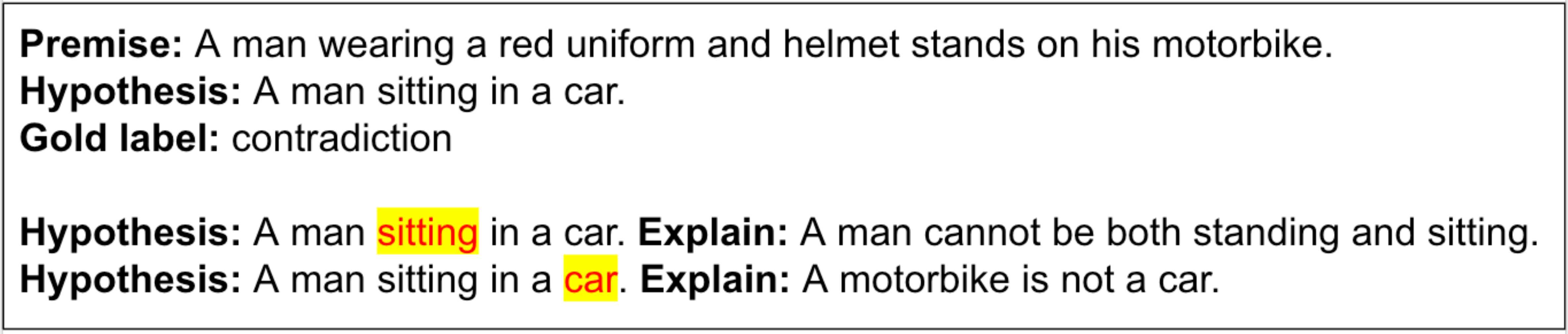}
         \caption{Example of different NL explanations using different rationales.}
         \label{fig:example-a}
     \end{subfigure}
     \hfill
     \begin{subfigure}[b]{0.5\textwidth}
         \centering
         \includegraphics[width=\textwidth]{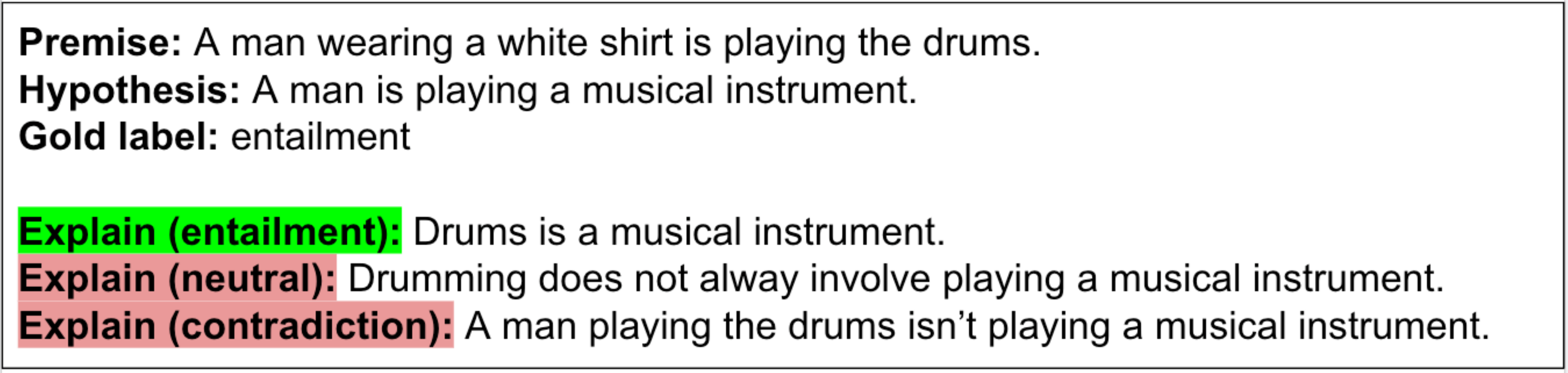}
         \caption{Example of correct and incorrect NL explanations generated for a same premise-hypothesis, but favoring different labels.}
         \label{fig:example-b}
     \end{subfigure}
        \caption{Examples of NLEs in NLI task.}
        \label{fig:exmples}
\end{figure}

\paragraph{Issue 1: Lack of rationale in NLE Generation}
\label{sec:issue-lack}

Current approaches for explanation generation produce only one specific explanation for each data instance. However, these approaches ignore the variability in human reasoning and alternative explanations. Annotators could assign the same label to a data instance by considering different rationales. For example, given the premise and the hypothesis shown in Figure~\ref{fig:example-a}, it is easy to infer that the label should be \textit{contradiction}. However, this label could be explained using two different rationales -- indicated by \textit{``sitting''} or by \textit{``car''}. Ignoring this aspect would limit the application of NLE in NLI tasks because trustworthy explanations should be consistent with the appropriate rationale used by humans to interpret the label. 

\paragraph{Issue 2: Inclusion of Spurious Explanations}
\label{sec:issue-incorrect}

NLEs that are inconsistent with commonsense logic provide little help for model prediction~\cite{camburu2020make}. For example, given the premise-hypothesis pair in Figure~\ref{fig:example-b}, the explanation regarding the correct label (\textit{entailment}) aligns with the fact that drums are indeed musical instruments. However, while tailored explanations could be generated to favor other labels -- \textit{neutral} and \textit{contradiction}, such explanations are themselves factually incorrect. This happens because while deep learning-based text generators are powerful enough to generate readable sentences, they often lack commonsense reasoning ability~\cite{zhou2020evaluating}. When generating an explanation, such models are prone to output negating text if conditioned to the \textit{contradiction} label, or output text with uncertainty if conditioned to the \textit{neutral} label, without reasoning about the plausibility of the generated text.

\paragraph{Proposed Solution: Language Inference with Relevant Explanations}

To address the aforementioned issues observed in how NLEs have been incorporated in NLI models, we propose a novel framework for Language Inference with Relevant Explanations (LIREx). LIREx augments the NLI model with relevant, plausible NLEs produced and selected by a rationale-enabled explanation generator and an instance selector. We conduct detailed analysis to show not only that our model is able to bring significant performance improvement to the NLI task, but also that the generated explanations are highly aligned with human interpretation when evaluated on a relevance-based evaluation metric.

\section{Related Work}
\label{sec:related-work}
NLEs have been studied along two main directions. The first direction focused on how explanations could be treated as contextual features to improve the model performance. Studies in this direction include~\cite{li2016learning,srivastava2017joint,wang2017naturalizing,hancock2018training,qin2020learning,lee2020lean}, in which the authors used semantic parsers to convert unstructured NLEs into structured feature-like logical forms. These logical forms could further benefit low-resource setting by weakly labeling more unlabeled data. One limitation of such approaches is that the localized contexts usually have limited ability to represent the semantic meaning of text and are often difficult to convert to logical forms when they get too complicated. 

The second direction focused on training NLE generators to justify model predictions, usually as a post-hoc exercise. \cite{kim2018textual} trained textual explanation generators conditioned on the video frames and commands in self-driving cars to describe and justify the operated actions. \cite{huk2018multimodal} proposed an explanation module for visual question answering and an activity recognition task, in which they first use encoders to jointly predict labels and infer the rationale regions of an image, and then generate text explanations by conditioning on the predicted labels and inferred rationales. \cite{camburu2018snli} suggested the inclusion of NLEs for NLI task by proposing e-SNLI, an expanded dataset that contains NLE annotations. They also jointly trained the prediction model and the explanation generation model conditioned on the predicted label. However, jointly training the two models led to a non-negligible loss in performance (by about 2 points in F1).

In recent studies, generated NLEs have been combined with original data for label prediction tasks. \cite{rajani2019explain} proposed CAGE for commonsense reasoning task, where they first trained an explanation generator to predict explanations based on the question and answer choices, and then expanded the original classifier input by combining questions and generated explanations. This strategy achieved a significant improvement over the baseline that only used the original data as input. \cite{kumar2020nile} attempted a similar approach, NILE, for the NLI task on the e-SNLI dataset but with a modification where, instead of generating one explanation per training instance, they trained three independent generators conditioned on each label (\textit{entailment}, \textit{neutral}, and \textit{contradiction}), respectively. Then the final NLI model takes as input, the premise-hypothesis pair as well as all three generated explanations. They also evaluated the faithfulness of the explanations to demonstrate that the explanations are well correlated with model predictions, but reported a drop in performance on the NLI task.






\section{Augmenting NLI with Relevant Explanations} \label{sec:approach}

\begin{figure*}
     \centering
     \includegraphics[width=0.9\textwidth]{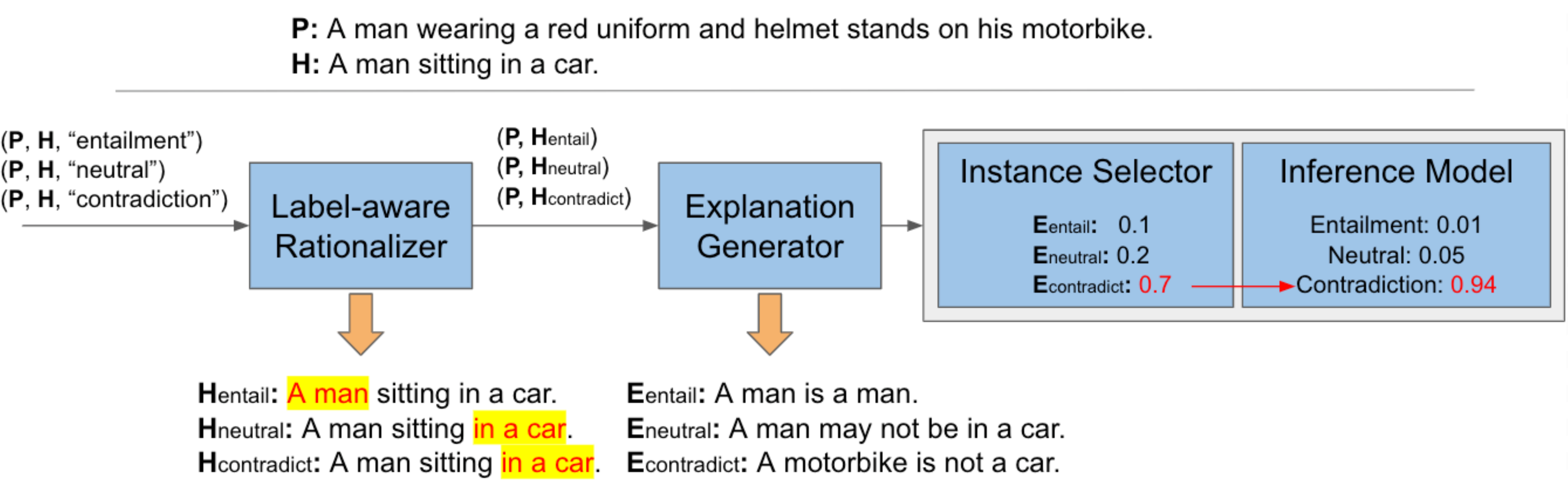}
        \caption{The overall workflow of LIREx framework}
        \label{fig:model-workflow}
\end{figure*}

The overall workflow of LIREx is shown in Figure~\ref{fig:model-workflow}. Given a premise-hypothesis (P-H) pair, a label-aware rationalizer predicts rationales by taking as input a triplet (P, H, $x$; $x\in \text{\{entail, neutral,  contradict\}}$) and outputs a rationalized P-H pair, (P, H$_{\text{x}}$). Next, the NLE generator generates explanations (E$_{\text{x}}$) for each rationalized P-H pair. Then, the explanations are combined with the original P-H pair as input to the instance selector and inference model to predict the final label. Each component is described below.

\subsection{Label-aware Rationalizer - R($\cdot$)}
\label{sec:rationalizer}

As described in~\cite{camburu2018snli}, NLEs are created based on the rationales highlighted by annotators. We simulate this process by using a rationalizer to provide the relevant rationales for the NLE generator. This operation models how the explanations are generated for human interpretation. 

We formulate this step as a token-level binary classification task where $1$ indicates a rationale token and $0$ indicates a background token. The rationale classification is only performed on the hypothesis because we consider the premise as background context for the NLI task, where the task is to justify if the hypothesis is an entailment, contradiction, or neutral statement with respect to the premise. Therefore, we hypothesize that the rationales in the hypothesis are sufficient to predict the correct label. We first construct the input sequence as $S^p$=\textit{$<$s$>$Label$<$s$>$Premise$<$s$>$} and $S^h$= \textit{$<$s$>$Hypothesis$<$s$>$}, where $<$s$>$ is a special token that separates the components.\footnote{RoBERTa includes two special tokens, $<$s$>$ and $<$/s$>$. For simplicity, we use $<$s$>$ to denote both of them.} We append the label information to the premise to inform the rationalizer to highlight label-related rationales. Then we use a RoBERTa$_{base}$ model~\cite{liu2019roberta} to extract hidden representations for $S^p$ and $S^h$, denoted as $\mat{H}^p=[..., \vec{h}_i^p, ...]$ and $\mat{H}^h=[..., \vec{h}_j^h, ...]$, respectively. Since the rationales in hypothesis depends on the semantic meaning of the premise, we use cross attention to embed premise into hypothesis, defined as
\begin{align}
\label{eq1:rationalier}
a_{ij} &= \frac{\text{exp}(\vec{h}_i^{hT}\text{Tanh}(\mat{W}_1^T\vec{h}_j^p))}{\sum_{k=0}^{L^p}\text{exp}(\vec{h}_i^{hT}\text{Tanh}(\mat{W}_1^T\vec{h}_k^p))}\\
\vec{\hat{h}_i^{h}} &= \text{concat}(\vec{h}_i^{h}, \text{Pool}(\mat{H}^p), \sum_ka_{i,k}\vec{h}_j^p)
\end{align}
where $a_{ij}$ denotes the attention score of the $j^{\rm th}$ token in $S^p$ to the $i^{\rm th}$ token in $S^h$, $L^p$ denotes the sequence length of $S^p$, and $\mat{W}_1$ is trainable parameter matrix. Then the new representation of the $i$th token in $S^h$ is created by concatenating its original state representation, maxpooling representation over $\mat{H}^p$, and the corresponding sum of attentional representation from $\mat{H}^p$. At last, we use a softmax layer with a linear transformation to model the probability of the $i^{\rm th}$ token in $S^h$ being a rationale token: $\text{P}(y_i^h|S^p, S^h)=\text{softmax}(\mat{W}_2\vec{\hat{h}}_i^{h})$, where $\mat{W}_2$ is a trainable parameter matrix for linear transformation on $\vec{\hat{h}}_i^{h}$.

\subsection{NLE Generator - G($\cdot$)}
\label{sec:generator}

We model NLE generation as a text generation task, in which we leverage GPT2~\cite{radford2019language}, a language model trained on large-scale language corpus. We choose GPT2$_{medium}$ so that we could have an end-to-end comparison with the previous study that uses the same architecture. 

In the previous study~\cite{kumar2020nile}, the authors finetuned GPT2 independently for each label. Specifically, they trained three GPT2 models separately, 
which are $\text{G}_{x}(\text{P},\text{H},\text{E})$, $x\in \text{\{entail, neutral,  contradict\}}$.
Each $\text{G}_x$ is trained only with the P-H pairs annotated as $x$. As described in the Introduction section, such setup is (a)~insensitive to the variety of human interpretation toward data, and (b)~results in spurious explanations that further harm the label inference task. Additionally, this generation strategy requires training $n$ GPT2 models for $n$ labels, which is still expensive even with fine-tuning. 
To solve the above issues, we train a single GPT2 model, $\text{G}(\text{P},\text{H}^*,\text{E})$, where $\text{H}^*$ is rationalized hypothesis. For example, for the P-H pair in Figure~\ref{fig:example-b}, we construct the input sequence, $S^g$, as:
\noindent\fbox{%
    \parbox{\columnwidth}{%
        Premise: \texttt{P} \\
        Hypothesis: A man is [playing] a [musical] [instrument].\\
        Explanation: \texttt{E}
    }%
}
\noindent where \texttt{P} and \texttt{E} represent the premise and NLE text in training data. To inform the generator about the rationales in hypothesis, we highlight rationale tokens by surrounding them with the square brackets `[]'. The generator is fine-tuned by modeling the text input as a whole. To generate an NLE, we simply remove \texttt{E} from $S^g$ and then use the rest as a text input prompt for the generator. 

Unlike the approach in \cite{kumar2020nile} where the label information is appended to the text, we hide the label from the generator to force the model to generate rationale-enabled NLEs. This is consistent with our goal to simulate diverse human interpretation, and prevents the model from generating spurious label-based explanations.

During training and evaluating the explanation generator, we use the rationale tokens and NLEs provided by human annotators. After the generator is trained, we generate three new NLEs for each instance based on the rationales by including each label in $S^p$, independently. Now each P-H pair is provided with three explanations and we remove the original gold explanations in training data. This is to prevent the instance selector model in the next step from overfitting on the training examples. 


\subsection{Instance Selector and Inference - $\text{S}(\cdot)$ and $\text{Infer}(\cdot)$}
\label{sec:selector-and-infer}

When a P-H pair and the generated explanations are fed into an inference model, the model benefits from the addition of the explanations when they are correct (cf.~Figure~\ref{fig:example-b}). On the other hand, incorrect explanations lead to large uncertainty during the inference process. So, we first select a single plausible explanation for the final inference. To achieve this, we develop a simple strategy assuming that when the labels are correct, the NLEs generated based on the corresponding enabled rationales are the correct explanations. This allows us to only estimate which NLE is generated by the gold label-enabled rationale. To further simplify this task, if we assume that the gold label-enabled explanation is more likely to be plausible than the other two explanations, we could identify the gold label-enabled explanation by accurately predicting the correct label for the standard NLI task. 
In other words, a good prediction on the NLE selection task can be achieved by just training a standard NLI classification model.

\paragraph{Training instance selector model} We initialize the selector \textbf{S($\cdot$)} with a RoBERTa$_{\rm base}$ model and use the representation of the first token, $\vec{h}_0$, as the sequence representation. On top of this, an output layer of linear transformation and activation, $\text{Tanh}(\mat{U}_1\vec{h}_0)\mat{U}_2$, is applied for prediction. $\mat{U}_1$ and $\mat{U}_2$ are parameter matrices.
We train \textbf{S($\cdot$)} as a standard supervised learning task where premise and hypothesis are concatenated as a single input sequence, and the model is trained to predict label $Y\in \{\text{entailment, neutral, contradiction\}}$. We pre-train the instance selector and use the label prediction probability distribution as the estimator to find the most likely explanation corresponding to the true label. 
To improve model robustness, during training, we sample the candidate explanations based on the probability distribution, instead of picking just the most likely explanation. This allows the inference model to better tolerate less plausible explanations. During test phase, we select the explanation with the highest probability.

\paragraph{Training inference model} Once the explanation instance is selected, we train the inference model, \textbf{Infer}(premise, hypothesis, explanation), with the same model architecture as the selector. 
Taking insights from the field of weak supervision~\cite{fries2017swellshark,bach2017learning,ratner2020snorkel} where weakly labeled data is used for training models,
we treat the selected explanations as weakly selected instances. Instead of using the standard cross-entropy loss (that requires gold label) as training objective, we use a probability-oriented training objective, soft cross-entropy loss, to improve the model robustness towards noisy input:
\begin{equation}
\label{eq3:soft-crossentropy}
\text{CE}_{soft}(\vec{p}, \vec{\hat{p}}) = \sum_{l\in{\text{\{e,n,c\}}}}\vec{\hat{p}}_{l} \log \vec{p}_l
\end{equation}
where $\vec{p}$ and $\vec{\hat{p}}$ are predicted and the ``target'' probabilities for each label, respectively. In our experiments, we use the estimated probabilities from the instance selector as our ``target'' probability for training inference model.

\section{Experiments and Results}

\subsection{Data sets}
The proposed framework is evaluated on two widely-used corpora for language inference -- SNLI~\cite{bowman2015large} and MultiNLI~\cite{williams2017broad}. 

SNLI is a balanced collection of P-H annotated pairs with labels from \{\textit{entailment}, \textit{neutral}, \textit{contradiction}\}. It consists of about $550$K, $10$K and $10$K examples for train, development, and test set, respectively. \cite{camburu2018snli} recently expanded this data set to e-SNLI in which each data instance is also annotated with explanations. Based on the gold labels for each P-H pair, annotators were asked to highlight the rationale tokens, and provide NLEs based on the rationale. Previous studies~\cite{camburu2018snli,kumar2020nile} removed over $17$K non-informative training examples (where the explanations contain the entire premise or hypothesis) from their analysis. In our work, to maintain the original training data with minimal changes, we hold out these non-informative training instances only when training the explanation generator, but use the full training data for the remaining steps.

The MultiNLI data set differs from the SNLI data set in that it covers a range of genres of spoken and written text. It contains $433$K P-H pairs annotated the same way as SNLI. The evaluation set is divided into Dev-match set ($10$K) and Dev-mismatch set ($10$K) -- the former is derived from same five domains as the training data and the latter is derived from five other domains.   


\subsection{Model Implementation Details}

We compare our results against NILE~\cite{kumar2020nile}. All NILE models are denoted as NILE$_{\rm model}$ reported by~\cite{kumar2020nile} and we designed LIREx models, LIREx$_{\rm model}$, accordingly to provide a fair head-to-head comparison. 

\textbf{NILE}$_{\rm base}$ and \textbf{LIREx}$_{\rm base}$: The baseline RoBERTa models that takes input only with the P-H pair. \textbf{LIREx}$_{\rm base}$ is the reproduced version made for fair comparison.

\textbf{NILE}$_{\rm expl}$ and \textbf{LIREx}$_{\rm expl\_*}$: The baseline models that take as input only the explanations. The difference is that LIREx$_{\rm expl\_*}$ uses the single explanation selected by the instance selector, while \textbf{NILE}$_{\rm expl}$ uses all three explanations.

\textbf{NILE}$_{\rm all}$: The base model that uses the P-H pair as well as all explanations as input by concatenating them into one input sequence, e.g. ``p$\small{<}$s$\small{>}$h$\small{<}$s$\small{>}$e$_1\small{<}$s$\small{>}$e$_2\small{<}$s$\small{>}$e$_3$'', where all components are separated by a special token $\small{<}$s$\small{>}$. 

\textbf{NILE}$_{\rm all\_extra}$: Same as NILE$_{\rm all}$, except that extra negative samples are created for valid (P, H, E) triplets by sampling from other explanations. 

\textbf{LIREx}$_{\rm all\_max}$ and \textbf{LIREx}$_{\rm all\_prob}$: These two models use the pre-trained instance selector to first select an explanation candidate, and then concatenate the explanation to the P-H pair for input to the inference model. LIREx$_{\rm all\_max}$ selects an explanation with the highest probability while LIREx$_{\rm all\_prob}$ samples the candidate based on the probability distribution.

\subsection{Results on In-domain Evaluation}

The model performance on SNLI data is summarized in Table~\ref{tab:eval-snli}. We re-implemented the NILE baseline (NILE$_{\rm base}$) as LIREx$_{\rm base}$ and achieved slight improvement (of $0.06$) over published results. The rest of the table shows model improvements compared to the baselines accordingly, with the relative improvements against the baselines summarized in the ``vs.baseline'' column.
\begin{table}[]
\small
\begin{tabular}{lp{12mm}p{10mm}cp{6mm}}\hline
\textbf{Model} & \textbf{Dev}& \textbf{Test} & \textbf{vs.baseline} & \textbf{+ Data} \\\hline
SemBERT$_{\rm large}$     & 92.0  & 91.6  & -    & no                   \\
SemBERT$_{\rm wwm}$       & 92.2  & 91.9  & -    & no                   \\\hline
NILE$_{\rm base}$      & 91.86 & 91.49 & -  & no                   \\
NILE$_{\rm expl}$     & 88.49 & 88.11 & -3.38  & no                   \\
NILE$_{\rm all}$     & 91.74 & 91.12 & -0.37   & no                   \\
NILE$_{\rm all\_extra}$ & 91.29 & 90.73 & -0.76  & yes                  \\ \hline
LIREx$_{\rm base}$      & 92.15{\tiny$\pm$.05} & 91.55{\tiny$\pm$.04} & -  & no                   \\
LIREx$_{\rm expl\_max}$      & 89.95{\tiny$\pm$.05}     & 89.73{\tiny$\pm$.04}     & -1.82   & no                   \\
LIREx$_{\rm expl\_prob}$      & 90.10{\tiny$\pm$.05}     & 90.03{\tiny$\pm$.05}     & -1.52   & no                   \\
LIREx$_{\rm all\_max}$ & 92.15{\tiny$\pm$.04} & 91.73{\tiny$\pm$.03} & +0.18 & no \\
LIREx$_{\rm all\_prob}$ & \textbf{92.22{\tiny$\pm$.03}} & \textbf{91.87{\tiny$\pm$.03}} & \textbf{+0.32*} & no  \\ \hline 
\end{tabular}
\caption{Accuracy performance of LIREx on SNLI data (average of five random runs). ``*'' denotes that the best model is statistically significant (at significance level of $0.05$ against baseline and $0.01$ against NILE).  ``+ Data'' denotes if additional training data was created. SemBERT~\cite{zhang2019semantics} is included to refer to the best-reported performance.}
\label{tab:eval-snli}
\end{table}

As shown in the table, our final model (LIREx$_{\rm all\_prob}$), is able to achieve an absolute performance gain of $0.32$ accuracy points. In addition, we also provide other variants of our model, namely, LIREx$_{\rm expl\_max}$, LIREx$_{\rm expl\_prob}$ and LIREx$_{\rm all\_max}$. All the provided models show that models that use the instance selector achieve better performance. Further, the sampling-based selection strategy (as in LIREx$_{\rm *\_prob}$) performs better than the greedy selection using highest value (as in LIREx$_{\rm *\_max}$).

\subsection{Results on Out-of-Domain Transfer Evaluation}

To test how well our model can generalize to a different data set, we directly apply our model trained on SNLI to an out-of-domain data set, MultiNLI. As shown in Table~\ref{tab:eval-multinli}, without any fine-tuning, our model achieves significantly better performance compared to NILE. When compared to the corresponding baselines, our model performance dropped by $0.27$ on dev-matched and improved slightly by $0.06$ on dev-mismatched. Performance of NILE models, in contrast, dropped significantly on MultiNLI. In addition, compared to the baseline, LIREx$_{\rm all\_prob}$, our final model performs similarly between dev-matched and dev-mismatched, indicating that the inclusion of explanations as supervision improves the overall generalizability of the model.
\begin{table}[]
\small
\begin{tabular}{p{20mm}p{6mm}cp{6mm}c}\hline
\multirow{2}{*}{\textbf{Model}} & \multicolumn{2}{l}{\textbf{Dev-Matched}} & \multicolumn{2}{l}{\textbf{Dev-MisMatched}} \\\cline{2-5}
      & \textbf{Acc}     & \textbf{vs.baseline}   & \textbf{Acc}   & \textbf{vs.baseline}   \\\hline
NILE$_{\rm base}$ & 79.29 & -  & 79.29 & - \\
NILE$_{\rm expl}$ & 61.33 & -17.96    & 61.98   & -17.31    \\
NILE$_{\rm all}$     & 77.07  & -2.22   & 77.22  & -2.07    \\
NILE$_{\rm all\_extra}$ & 72.91 & -6.38 & 73.04  & -6.25   \\ \hline
LIREx$_{\rm base}$    & \textbf{80.12}  & -   & 79.73   & - \\
LIREx$_{\rm expl\_max}$   & 65.53 & -17.59   & 65.19  & -14.54    \\
LIREx$_{\rm expl\_prob}$   & 65.57 & -17.63   & 65.32  & -14.68    \\
LIREx$_{\rm all\_max}$    & 79.71   & -0.41   & 79.50     & -0.23       \\
LIREx$_{\rm all\_prob}$ & 79.85    & \textbf{-0.27}  & \textbf{79.79}   & \textbf{+0.06} \\\hline         
\end{tabular}
\caption{Transfer performance of LIREx on the out-of-domain MultiNLI data (average of five random runs)  without fine-tuning.}
\label{tab:eval-multinli}
\end{table}

\section{Discussion}
\subsection{Behavior of Rationalizer}

As described earlier, the explanations are generated based on rationales in the hypothesis. Heuristically, we could evaluate the rationalizer simply by looking at the F1 score to see how well the predictions match the true rationales. 
However, this evaluation strategy alone is insufficient, because annotators may include additional neighboring tokens as rationales. For example, for the hypothesis ``A man sitting in a car'' in Figure~\ref{fig:example-a}, ``sitting in a car'', ``in a car'', ``a car'', and ``car'' are all reasonably correct rationales as they all contain the most important rationale token ``car''. If an annotator provides
``in a car'' as rationale and the rationalizer predicts only ``car'', the automated F1 metric will be low even though the main rationale was correctly identified. So, in addition to the F1-based evaluation, we also conducted a manual verification of $100$ randomly sampled test examples, and report the instance-level accuracy in Table~\ref{tab:eval-rationale}. The manual verification was conducted by two annotators. The annotators were presented with P-H pairs, predicted rationales, and gold rationales, and were asked a ``Yes/No'' question: Do predicted rationales contain the key information from the gold rationales? Examples annotated as ``Yes'' were treated as correct predictions. 
The difference between automated and human evaluation shows that, although the rationalizer did not identify the exact human-provided rationales, it did identify the most important rationales (e.g., ``car'') with high accuracy.

\begin{table}[]
\small
\begin{tabular}{p{11mm}p{5mm}p{5mm}p{5mm}p{5mm}p{5mm}p{6mm}p{8mm}}\hline
\multirow{2}{*}{\textbf{Model}} & \multicolumn{3}{c}{\textbf{Dev}}   & \multicolumn{3}{c}{\textbf{Test}} & \textbf{Human} \\\cline{2-7}
                      & \multicolumn{1}{c}{\textbf{P}} & \multicolumn{1}{c}{\textbf{R}} & \multicolumn{1}{c}{\textbf{F1}} & \multicolumn{1}{c}{\textbf{P}} & \multicolumn{1}{c}{\textbf{R}} & \multicolumn{1}{c}{\textbf{F1}} & 
                      \multicolumn{1}{c}{\textbf{Eval}} 
                      \\\hline
Rationale & 59.40     & 65.22  & 62.17 & 59.21 & 64.89  & 61.92 & 90.53\\ \hline
\end{tabular}
\caption{F1-based performance of the rationalizer and instance-level accuracy of human evaluation from two annotators over $100$ randomly sampled test examples with an inter-rater agreement of 0.89.}
\label{tab:eval-rationale}
\end{table}

An \textbf{unexpected, yet preferred} behavior of the rationalizer is that, when there are no obvious rationales towards the pre-appended label information, the model tends to identify rationales that relate to the correct label. For example, for the hypothesis (a) in Table~\ref{tab:example-rationale}, we devised three ``alternate'' hypotheses to analyze how the rationale predictions change with different hypotheses.
We progressively modified a specific component in the original hypothesis -- e.g.~added \textit{``on stage''} in (b), replaced \textit{``man''} with \textit{``woman''} in (c), and included both these changes and replaced \textit{``musical instrument''} with \textit{``guitar''} in (d). We found that when rationales of a particular label is absent, the rationalizer is prone to output rationales of the correct label (as contradiction rationales in hypothesis (a) and neutral rationales in hypothesis (c)). 

Furthermore, when rationales of more than one label exist, the model is capable of identifying at least some correct rationales regarding a label. For example in Hypothesis (b), ``playing musical instrument'' is an entailment-related rationale, while ``playing musical instrument on stage'' is a neutral-related rationale. For hypothesis (d), we are able to correctly catch rationales of all labels (``playing'' is an entailment, ``stage'' is neutral and ``guitar'' is a contradiction). 

\begin{table}[!t]
\small
\begin{tabular}{p{25mm}p{6mm}l}
\hline
\multicolumn{3}{l}{\textbf{Premise:} A man wearing shirt is playing the drums.}\\\hline
\textbf{Hypothesis} & \textbf{Label} & \textbf{Rationales}   \\\hline
\multirow{3}{*}{\begin{tabular}[c]{@{}l@{}}(a) A man is\\playing a musical\\ instrument\end{tabular}}  & \colorbox{green}{E} & musical instrument  \\
                           & \colorbox{white}{N}  & musical instrument \\
                            & \colorbox{white}{C}  & musical instrument \\\hline
\multirow{3}{*}{\begin{tabular}[c]{@{}l@{}}(b) A man is\\playing a musical\\instrument on stage\end{tabular}} & \colorbox{white}{E}   & playing, musical instrument  \\
                     & \colorbox{green}{N}      & musical instrument, stage \\
                     & \colorbox{white}{C}       & playing, musical instrument  \\\hline
\multirow{3}{*}{\begin{tabular}[c]{@{}l@{}}(c) A woman is\\playing a musical\\instrument\end{tabular}}    & \colorbox{white}{E}  & musical instrument \\
                             & \colorbox{white}{N}   & woman, musical instrument \\
                             & \colorbox{green}{C}    & woman, musical instrument \\\hline
\multirow{3}{*}{\begin{tabular}[c]{@{}l@{}}(d) A woman is\\playing guitar on\\stage\end{tabular}}     & \colorbox{white}{E}  & playing \\
                        & \colorbox{white}{N}   & woman, stage \\
                         & \colorbox{green}{C}   & woman, guitar \\\hline              
\end{tabular}
\caption{Examples of the rationalizer behavior. All hypotheses share the given premise. Rationales are predicted by appending a corresponding label information to the premise. The true labels of the P-H pairs are highlighted in green.}
\label{tab:example-rationale}
\end{table}

\subsection{Analysis of Explanation Generator}
We conducted detailed analyses to show 
(a)~why label information should be removed from generation prompts and (b)~robustness of the generator towards rationale variants.

Removing label information from generation prompts prevents the generator from producing too many spurious explanations. In Table~\ref{tab:example-generation}, we present an example to compare the explanations generated when label information is either included or excluded. For the provided P-H, all three models use GPT2 to generate explanations for each label. \textbf{NILE} uses the plain premise and hypothesis combined with label information as generation prompts. \textbf{LIREx-w-label} uses the same information as \textbf{NILE}, and also includes an indication of rationale tokens in the format described in the NLE Generator section. Finally, in \textbf{LIREx-w/o-label}, the label information is removed. As shown in the example, all three models are able to generate reasonable explanations. However, when label information in appended to the input data, spurious explanations are presented (e.g.~for E and N labels in \textbf{NILE}, and the E label in \textbf{LIREx-w-label}). This is because the addition of label serves as a very strong constraint for the generator, which then leads to an explanation that is constructed using the label information as guidance. Even with rationales are provided, the issue could only be slightly alleviated (eg.~for N label in \textbf{LIREx-w-label}). 

However, the issue is resolved when labels are excluded from the generator because the generator is not forced to generate a label-aligned explanation, which could be spurious. Instead, it utilizes only the rationales and the original context to construct the explanations. As discussed earlier, when the rationale of a specified label does not exist, the rationalizer tends to provide rationales related to the true label. So we obtain three reasonable explanations for the E, N, and C labels in \textbf{LIREx-w/o-label}(E,N,C). Further, this would greatly increase the chance of a plausible explanation being selected as an explanation candidate by the instance selector.

\begin{table}[h]
\small
\begin{tabular}{p{2mm}p{22mm}p{22mm}l}\hline
\multicolumn{4}{l}{\begin{tabular}[c]{@{}l@{}}\textbf{Premise:}  Families waiting in line at an amusement park\\for their turn to ride.\\ \textbf{Hypothesis:} People are waiting to see a movie.\end{tabular}}  \\\hline
\textbf{L} & \textbf{NILE} & \textbf{LIREx-w-label} & \textbf{LIREx-w/o-label} \\\hline
& & \multicolumn{2}{l}{\textbf{Rationales:} people, waiting, see, movie} \\\cline{3-4}
E 
& \begin{tabular}[c]{@{}l@{}}People are\\waiting to see\\a movie at an\\amusement park.\end{tabular}                              
& \begin{tabular}[c]{@{}l@{}}Families are\\people. If\\people are\\waiting in line,\\it means that\\they are waiting\\to see a movie.\end{tabular} 
& \begin{tabular}[c]{@{}l@{}}People waiting\\in line are not\\waiting to see\\a movie.\end{tabular}  \\\hline
& & \multicolumn{2}{l}{\textbf{Rationales:} movie} \\\cline{3-4}
\colorbox{white}{N}
& \begin{tabular}[c]{@{}l@{}}Not all rides at\\amusement parks\\are a movie.\end{tabular}  & \begin{tabular}[c]{@{}l@{}}You cannot infer\\it's for a movie.\end{tabular}
& \begin{tabular}[c]{@{}l@{}}Amusement\\park and movie\\are not the same.\end{tabular}\\\hline
& & \multicolumn{2}{l}{\textbf{Rationales:} waiting, see, movie} \\\cline{3-4}
\colorbox{green}{C}
& \begin{tabular}[c]{@{}l@{}}People cannot be\\waiting to ride an\\amusement park \\and a movie \\at the same time.\end{tabular}
& \begin{tabular}[c]{@{}l@{}}People are either\\waiting to ride an\\amusement park,\\or waiting to see\\a movie.\end{tabular}
& \begin{tabular}[c]{@{}l@{}}People are either\\waiting in line\\for a ride or\\waiting to see\\a movie,\\but not both.\end{tabular}\\\hline
\end{tabular}
\caption{Example of NLEs generated when label information is included and excluded. For each label, we present the generated explanation from each system. The highlighted green ``C'' is the correct label for the P-H pair.}
\label{tab:example-generation}
\end{table}

\paragraph{Robustness of generator towards varied rationales}
The example in Table~\ref{tab:example-robust} shows different explanations generated by using one hypothesis token at-a-time as rationale. Out of the seven generated explanations, only two are invalid. In reality, however, it is rare to have only non-informative stopwords as rationales. This observation strengthens our argument that including rationales in the training of the explanation generator could improve the robustness of the model. 


\begin{table}[h]
\small
\begin{tabular}{p{10mm}p{50mm}c}\\\hline
\multicolumn{3}{l}{\begin{tabular}[c]{@{}l@{}}\textbf{Premise:}  Families waiting in line at an amusement park\\for their turn to ride.\\ \textbf{Hypothesis:} People are waiting to see a movie.\end{tabular}}  \\\hline
\textbf{Rationale} & \textbf{Explanation} & \textbf{Valid}\\\hline
People & Families implies more than one person & $\surd$\\
are & are waiting and waiting are different & $\times$\\
waiting & You can not infer they are waiting & $\times$ \\
to & Rides does not imply to see a movie & $\surd$ \\
see & To ride and to see are different & $\surd$\\
a & One cannot ride and see simultaneously. & $\surd$\\
movie & Just because families are waiting in line at amusement park doesn't mean they are waiting to see a movie & $\surd$\\\hline 
\end{tabular}
\caption{Example of explanations generated using different rationales -- one hypothesis token at-a-time as rationale.}
\label{tab:example-robust}
\end{table}

\subsection{Effect of Spurious Explanation}
As presented in Tables~\ref{tab:example-generation} and~\ref{tab:example-robust}, \textbf{LIREx} is able to consistently generate plausible NLEs with rationale-enabled explanations while \textbf{NILE} tends to generate spurious NLEs due to the inclusion of label information in explanation generation. To show how the spurious explanations could affect model performance, we train our model with the best NLE from the selector for each data instance, and then use a randomly selected NLE during evaluation. The results are presented in Table~\ref{tab:effect-spurious}. We observe that since \textbf{LIREx} is trained with rationale-enabled NLEs, it suffers only a small performance drop when presented with a randomly selected NLE. 
On the other hand, if we randomly select an NLE generated by \textbf{NILE}, the performance drops significantly compared to when choosing just the best NLE. This shows that (a)~\textbf{NILE} has a tendency to generate more spurious explanations, and (b)~if a spurious explanation is used for training the model, the performance drops significantly. On the other hand, \textbf{LIREx} does not use labels when training the generator, and hence, produces fewer spurious explanations, so even a randomly-selected explanation is still relevant.

\begin{table}[!t]
\small
\begin{tabular}{p{12mm}p{4mm}p{4mm}p{4mm}p{4mm}p{4mm}p{4mm}p{4mm}p{4mm}}\\\hline
\textbf{} & \multicolumn{2}{l}{\textbf{SNLI-Dev}} & \multicolumn{2}{l}{\textbf{SNLI-Test}} & 
\multicolumn{2}{l}{\textbf{MNLI-M}} & \multicolumn{2}{l}{\textbf{SNLI-Mis}}\\\hline
   & best & rand  & best  & rand & best & rand  & best  & rand\\\hline
\textbf{LIREx} & 92.15  & 91.84 & 91.80  & 91.55 & 79.51  & 79.51  & 79.52    & 79.33  \\
\textbf{LIREx$_{\rm NILE}$}   & 91.59   & 85.97   & 91.46 & 85.58 & 79.48  & 72.01  & 79.50    & 71.97 \\\hline         
\end{tabular}
\caption{Effects of spurious explanations on model performance. LIREx$_{\rm NILE}$ uses the same LIREx architecture but with the explanations generated from NILE.}
\label{tab:effect-spurious}
\end{table}

\subsection{Faithfulness Evaluation}
It is argued by~\cite{deyoung2019eraser} that a rationale-augmented classifier may not necessarily rely on the rationales but on the original data. Therefore, they propose to measure the faithfulness of the rationales by measuring the \textit{comprehensiveness} (removing rationales from input) and \textit{sufficiency} (using only the rationales as input). Since the \textbf{LIREx} inference model uses the generated explanations instead of rationales as input, following~\cite{kumar2020nile}, we probe the model by removing explanations and using just the explanations to measure comprehensiveness and sufficiency. As shown in Table~\ref{tab:probing}, when compared to the complete input, removal of explanation from the input reduces the performance on all data sets. Just using explanations leads to a significantly larger drop in performance, which is expected because an explanation is more meaningful when combined with the appropriate context (P-H pair) rather than by itself. These two observations show that the model depends on both the P-H pairs and explanations to make predictions, and that the explanations do demonstrate faithfulness. However, it is not clear why the effect on comprehensiveness is not as significant as that on sufficiency. 


\begin{table}[]
\small
\begin{tabular}{lcccc}\hline
\textbf{}    & \textbf{SNLI-Dev} & \textbf{SNLI-Test} & \textbf{MNLI-M} & \textbf{MNLI-Mis} \\\hline
D+E & 92.22 & 91.87    & 79.85  & 79.79  \\
D  & 90.95 & 91.07 & 77.10 & 76.88  \\
E & 62.40  & 62.21  & 43.35  & 43.93 \\\hline 
\end{tabular}
\caption{Faithfulness analysis of LIREx on both SNLI and MultiNLI data. ``D+E'' uses both P-H pairs and selected explanations as inputs for inference model, ``D'' uses only the P-H pair, and ``E'' uses only the selected explanations.}
\label{tab:probing}
\end{table}

\subsection{Relevance Evaluation}
Finally, we postulate that trustworthy explanations should be consistent with the appropriate rationale used to interpret the label. Given a specific example that contains different rationales leading a same label, the generator should be able to generate different yet reasonable explanations for each kind of rationale (cf.~Figure~\ref{fig:example-a}). We analyze the generated NLEs based on their  \textbf{relevance} to human interpretation. From each data set, we randomly sampled $100$ examples and ask two annotators to judge the relevance of the generated explanations. Each annotator was provided with context information (premise, hypothesis, rationale, and explanation), and asked to label them as $1$ if they agree that the information about the rationales is contained in the explanation, or $0$ otherwise. Since \textbf{NILE} does not use rationale to generate explanations, we use human-provided rationales in the dataset as the reference target. For \textbf{LIREx}, we used predicted rationales as reference targets. As shown in Table~\ref{tab:eval-relevance}, \textbf{LIREx} is able to maintain a high relevance score between explanations and predicted rationales, even when transferred to the out-of-domain data sets. This shows that the rationale-enabled explanations in \textbf{LIREx} are more aligned with human interpretation of the rationales.

\begin{table}[]
\small
\begin{tabular}{lcccc}\hline
\textbf{}     & \textbf{SNLI-Dev} & \textbf{SNLI-Test} & \textbf{MNLI-M} & \textbf{MNLI-Mis} \\\hline
NILE & 84 & 84 & - & -  \\
LIREx  & 99 & 97  & 95  & 95 \\\hline                      
\end{tabular}
\caption{Manual evaluation of the relevance score over $100$ randomly sampled data from each data set by two annotators with the inter-rater agreement of $0.95$. NILE evaluations of MultiNLI corpus are missing because we do not have ground truth rationales from human annotators.}
\label{tab:eval-relevance}
\end{table}

\section{Conclusion}
In this work, we identified two flaws in the current strategy of using NLEs for the NLI task. To overcome these limitations, we proposed a novel framework, called LIREx, that incorporates both a rationale-enabled explanation generator and an instance selector to augment NLI models with only relevant, plausible NLEs. The code is available at \url{https://github.com/zhaoxy92/LIREx}. The proposed framework achieves a significant improvement over a strong baseline by $0.32$ accuracy points on the SNLI data set, and is comparable to the current state-of-the-art performance on the task. When evaluated over an out-of-domain MultiNLI data set, the proposed approach demonstrated significantly better performance than previously published results without fine-tuning. We conducted extensive qualitative analysis to evaluate each component of our model. Qualitative analysis showed that LIREx generates flexible, faithful, and relevant NLEs that allow the model to be more robust to spurious explanations and better aligned to human interpretation. This work demonstrates the importance and usefulness of including human interpretation in NLI models.



\bibliography{references}

\section{Appendix}
\appendix

\section{A. Model Implementation and Training Details}

\paragraph{Parameter Matrices} 
Here we give the details of the parameter matrices introduced in rationalizer, instance selector, and inference model. 

In \textbf{Rationalizer:} The two parameter matrices, $\mat{W}_1$ in Eq.1 and $\mat{W}_2$ in softmax layer are randomly initialized with the dimensions of $768\times 768$ and $2\times 2304$, respectively. 

In \textbf{Instance Selector} and \textbf{Inference:} With the state representations of the RoBERTa$_{\rm base}$ model, we use the representation of the first token (denoted as $\vec{h}_0$) as the corresponding input sequence representation. Then linear transformations and the Tanh activation are applied via $\text{Tanh}(\mat{U}_1\vec{h}_0)\mat{U}_2$, where $\mat{U}_1$ and $\mat{U}_2$ are parameter matrices with the dimensions of $768\times 768$ and $768 \times 3$, respectively.

\paragraph{Training Hyper-parameters}. In Table~\ref{tab:hyper-parameter}, we present all hyper-parameters customized (encoder model, batch size, learning rate, and number of training epochs) for fine-tuning all the pre-trained models in LIREx. All other hyper-parameters (e.g. dropout rates, max sequence length, etc.) are kept the same with the default setting of the pre-trained encoders.

\begin{table}[h]
\small
\begin{tabular}{lcccc} \hline
    & \textbf{Encoder} & \textbf{Batch} & \textbf{Learning rate} &  \textbf{Epochs} \\\hline
R($\cdot$) & RoBERTa$_{\rm base}$  & 32 & 1e-5 & 10  \\
G($\cdot$) & GPT2$_{\rm medium}$& 1 & 2e-5 & 2  \\
S($\cdot$) & RoBERTa$_{\rm base}$ & 64 & 2e-5  & 3  \\
Infer($\cdot$)  & RoBERTa$_{\rm base}$ & 64  & 2e-5 & 3 \\\hline        
\end{tabular}
\caption{Training hyper-parameters for each LIREx component. R($\cdot$), G($\cdot$), S($\cdot$), and Infer($\cdot$) represent the rationalizer, NLE generator, instance selector, and inference model, respectively.}
\label{tab:hyper-parameter}
\end{table}

\paragraph{Software and Library}
Our code is implemented in Pytorch \url{http://pytorch.org} and the encoders are leveraged from the pre-trained models in \url{http://huggingface.co}.

\paragraph{Hard Device} All experiments are done on a single NVidia Tesla V100 GPU with 16GB memory. We also report that, for a smaller GPU (GeForce GTX 1080ti with 11GB memory), R($\cdot$), S($\cdot$), and Infer($\cdot$) can be trained without doing any modifications, and G($\cdot$) needs to be changed to GPT2$_{\rm small}$.

\section{B. Standard Deviations of LIREx Performance}
In Table~\ref{tab:std}, we present the standard deviations of the LIREx models used in the paper.

\begin{table}[h]
\center
\small
\begin{tabular}{lcc}\hline
 & \textbf{Mean}  & \textbf{Std Dev}  \\\hline
LIREx$_{\rm base}$      & 92.15 & 0.05 \\
LIREx$_{\rm expl\_max}$  & 89.95 & 0.05 \\
LIREx$_{\rm expl\_prob}$ & 90.10 & 0.05 \\
LIREx$_{\rm all\_max}$   & 92.15 & 0.04 \\
LIREx$_{\rm all\_prob}$  & 92.22 & 0.03\\\hline
\end{tabular}
\caption{Means and standard deviations of the LIREx performance on SNLI development set. The scores for each model are calculated based on five random runs.}
\label{tab:std}
\end{table}

\end{document}